\documentclass[conference]{IEEEtran}
\IEEEoverridecommandlockouts
\usepackage{xcolor}
\usepackage{colortbl}
\usepackage{graphicx}   
\usepackage{xcolor} 
\usepackage{cite}
\usepackage{amsmath,amssymb,amsfonts}
\usepackage{algorithmic}
\usepackage{graphicx}
\usepackage{textcomp}
\usepackage{xcolor}
\usepackage{amsmath}    
\usepackage{amssymb}    
\usepackage{bm} 
\def\BibTeX{{\rm B\kern-.05em{\sc i\kern-.025em b}\kern-.08em
    T\kern-.1667em\lower.7ex\hbox{E}\kern-.125emX}}

\begin{document}
\title{
    \includegraphics[height=1.0cm]{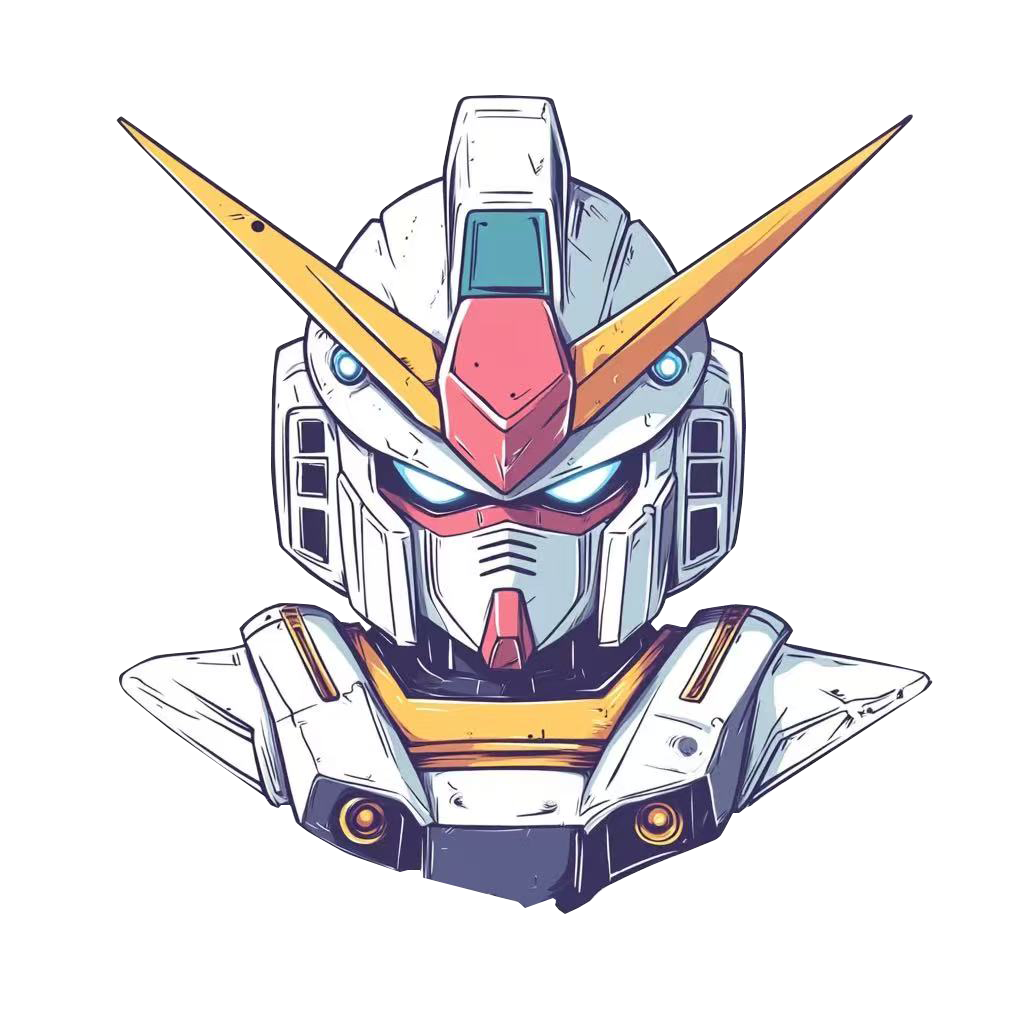}~ 
    \textcolor{gray}{Gundam}\textcolor{red}{Q}: Multi-Scale Spatio-Temporal Representation Learning for Robust Robot Path Planning \\

}


\author{
    Yutong Shen\textsuperscript{\rm 1}*,
    Ruizhe Xia\textsuperscript{\rm 1}*,
    Bokai Yan\textsuperscript{\rm 1},
    Shunqi Zhang\textsuperscript{\rm 1},
    Pengrui Xiang\textsuperscript{\rm 1},
    Sicheng He\textsuperscript{\rm 1},\\
    Yixin Xu\textsuperscript{\rm 3}\\
    *Corresponding author.
}

\maketitle

\begin{abstract}
In dynamic and uncertain environments, robotic path planning demands accurate spatiotemporal environment understanding combined with robust decision-making under partial observability. However, current deep reinforcement learning-based path planning methods face two fundamental limitations: (1) insufficient modeling of multi-scale temporal dependencies, resulting in suboptimal adaptability in dynamic scenarios, and (2) inefficient exploration-exploitation balance, leading to degraded path quality. To address these challenges, we propose GundamQ: A Multi-Scale Spatiotemporal Q-Network for Robotic Path Planning. The framework comprises two key modules: (i) the Spatiotemporal Perception module, which hierarchically extracts multi-granularity spatial features and multi-scale temporal dependencies ranging from instantaneous to extended time horizons, thereby improving perception accuracy in dynamic environments; and (ii) the Adaptive Policy Optimization module, which balances exploration and exploitation during training while optimizing for smoothness and collision probability through constrained policy updates. Experiments in dynamic environments demonstrate that GundamQ achieves a 15.3\% improvement in success rate and a 21.7\% increase in overall path quality, significantly outperforming existing state-of-the-art methods.

\end{abstract}

\section{Introduction}

Robot path planning in dynamic environments aims to design optimal trajectories for autonomous mobile robots from a start to a goal location. The task is challenging not only because of the complexity of adapting to dynamic environments,but also due to the requirement for precise spatiotemporal perception. The successful completion of such tasks in dynamic environments is critical for applications in autonomous driving, service robotics, and rescue operations.

Recent studies reveal that existing deep reinforcement learning (DRL)-based approaches exhibit limited adaptability in dynamic environments and require extensive training compared to traditional path planning methods. This performance gap significantly restricts their deployment in real robotic systems. Previous studies \cite{b1} have explored multi-robot path planning using diffusion models combined with classical search algorithms, human-in-the-loop feedback \cite{b2} for goal-conditioned exploration, attention-based DRL methods \cite{b3} for focusing on critical regions, and hybrid reinforcement learning with Rapidly-exploring Random Trees (RRT) \cite{b4}. However, these approaches either rely on single-scale models or fail to explicitly model multi-scale temporal dependencies, limiting adaptability in complex environments. Additionally, many methods emphasize local exploration and rely heavily on historical experience, resulting in suboptimal exploration-exploitation balance.
\begin{figure}[t]
    \centering
    \includegraphics[width=0.85\linewidth]{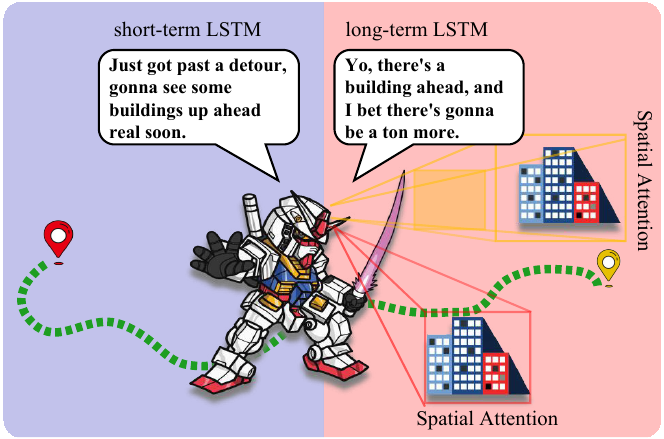} 
    \caption{This figure illustrates the perception mechanism for path planning. The spatial attention module, implemented with convolutional layers, highlights critical objects with high discriminative capability (shown in red). In addition, a multi-scale LSTM is employed to jointly capture past dependencies and predict future dynamics, thereby enhancing the effectiveness of path planning.}
    \label{fig:pdfimage}
\end{figure}

To overcome these limitations, we propose the \textbf{GundamQ framework}. Our system integrates a spatiotemporal perception module that combines multi-scale temporal modeling with spatial attention mechanisms to enhance environmental understanding and dynamics prediction. Simultaneously, an adaptive strategy optimization module intelligently balances exploration intensity and decision quality based on environmental dynamics and task requirements. The integration of these modules enables the generation of safe and efficient trajectories in complex dynamic scenarios while maintaining real-time performance.

Our contributions are summarized as follows:
\begin{itemize}
    \item We propose the \textbf{GundamQ framework}, which innovatively integrates a multi-scale LSTM network with spatial attention mechanisms, effectively addressing insufficient spatiotemporal feature extraction in dynamic environments.
    \item We design an \textbf{adaptive exploration-optimization mechanism} that dynamically balances exploration and exploitation based on environmental and task characteristics, significantly improving path planning performance.
    \item We develop a \textbf{hierarchical reward function and simulation platform}, integrating global navigation, local obstacle avoidance, and motion smoothness as multi-dimensional objectives, and demonstrate through experiments that our method outperforms state-of-the-art approaches, improving success rate by 15.3\% and path quality by 21.7\% in dynamic scenarios.
\end{itemize}

\section{Related works}
Our work focuses onDecision-Making in Dynamic Environments, taking robotic path planning as the primary task, and we propose a novel Adaptive Q-Network architecture to optimize navigation performance.

\subsection{Dynamic Environment Decision-Making }

Decision-making in dynamic environments refers to the process by which an agent makes optimal real-time decisions in time-varying environments. Existing approaches can be grouped into three categories. Graph-based prediction methods \cite{b5} construct dynamic graph structures online to enable long-horizon path planning, yet they struggle with multi-scale temporal dependencies and exploration–exploitation trade-offs. Hierarchical reinforcement learning methods \cite{b6} decompose decision-making into high- and low-level controllers to improve long-horizon performance, yet information transmission across levels may result in the loss of time-sensitive features. Model-based reinforcement learning methods construct dynamic environmental models and combine short-term prediction with online planning to balance safety and efficiency; however, their effectiveness depends heavily on model accuracy. Overall, these methods either focus solely on short-term dynamics or rely on fixed time windows, limiting adaptability in complex dynamic environments. Moreover, they often adopt static exploration strategies, reducing efficiency and leading to suboptimal path quality and slower convergence.

\subsection{Robot Path Planning}

Robot path planning aims to design collision-free optimal trajectories for autonomous mobile robots from start to goal positions. DRL-based path planning methods can be categorized into two main approaches. Value-based methods learn a state-action value function (Q-function) \cite{b8} to guide optimal path selection. For example, dueling network architectures improve learning stability and policy performance, but they struggle to capture long-term dependencies and adapt to dynamic environmental changes. Policy-gradient methods \cite{b10} directly optimize a policy to select actions. Distributed reinforcement learning has been applied for multi-robot cooperative exploration, yet these methods tend to over-exploit known strategies, limiting the discovery of new paths and potentially producing suboptimal trajectories. Overall, despite their advantages, current DRL-based path planning algorithms still face two critical challenges in complex dynamic environments: limited capability to jointly model long- and short-term behavior patterns, and insufficiently balanced exploration-exploitation strategies, which constrain path quality and robustness.

\subsection{Q-Networks}
Q-Networks guide agents to select optimal actions in complex environments by learning state-action value functions (Q-functions). The mainstream approaches can be categorized into three types: the standard Q-Network (DQN) \cite{b11} employs deep neural networks to approximate the Q-function, combined with experience replay and target networks, but may suffer from instability and slow convergence in high-dimensional state spaces; Double DQN \cite{b12} utilizes two separate networks for action selection and evaluation to reduce overestimation, yet strategy updates can be insufficient in high-dimensional continuous action spaces; Deep Deterministic Policy Gradient (DDPG) \cite{b13} is suitable for continuous action domains. Overall, these methods struggle to capture long-term temporal dependencies in dynamic environments and often fail to achieve effective exploration-exploitation balance, thereby limiting both path quality and planning efficiency.

\begin{figure*}[t]
    \centering
    \includegraphics[width=0.85\textwidth,page=1]{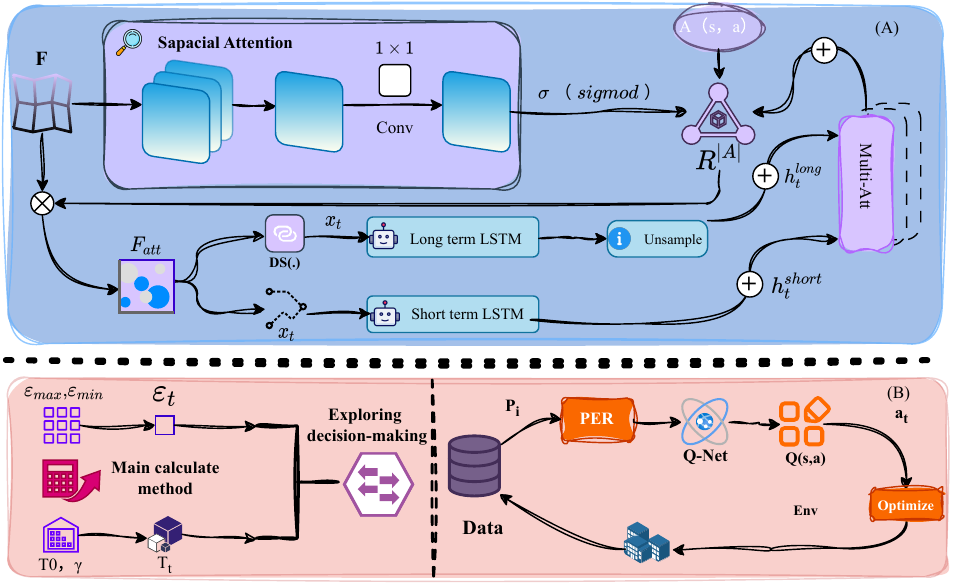}
    \caption{GundamQ framework.
(a) Perception module: Input feature 
F is processed with spatial attention (1×1 conv) and convolutions, then fed into short- and long-term LSTMs to capture local dynamics and long-range trends, followed by multi-branch fusion for spatiotemporal representation.
(b) Adaptive Policy Optimization module: Fused features are used by the decision unit and Q-NetPER to compute action values, which are updated adaptively with environmental feedback.}
    \label{fig:pdf}
\end{figure*}

\begin{figure*}[t]
    \centering
    \includegraphics[width=0.85\textwidth,page=1]{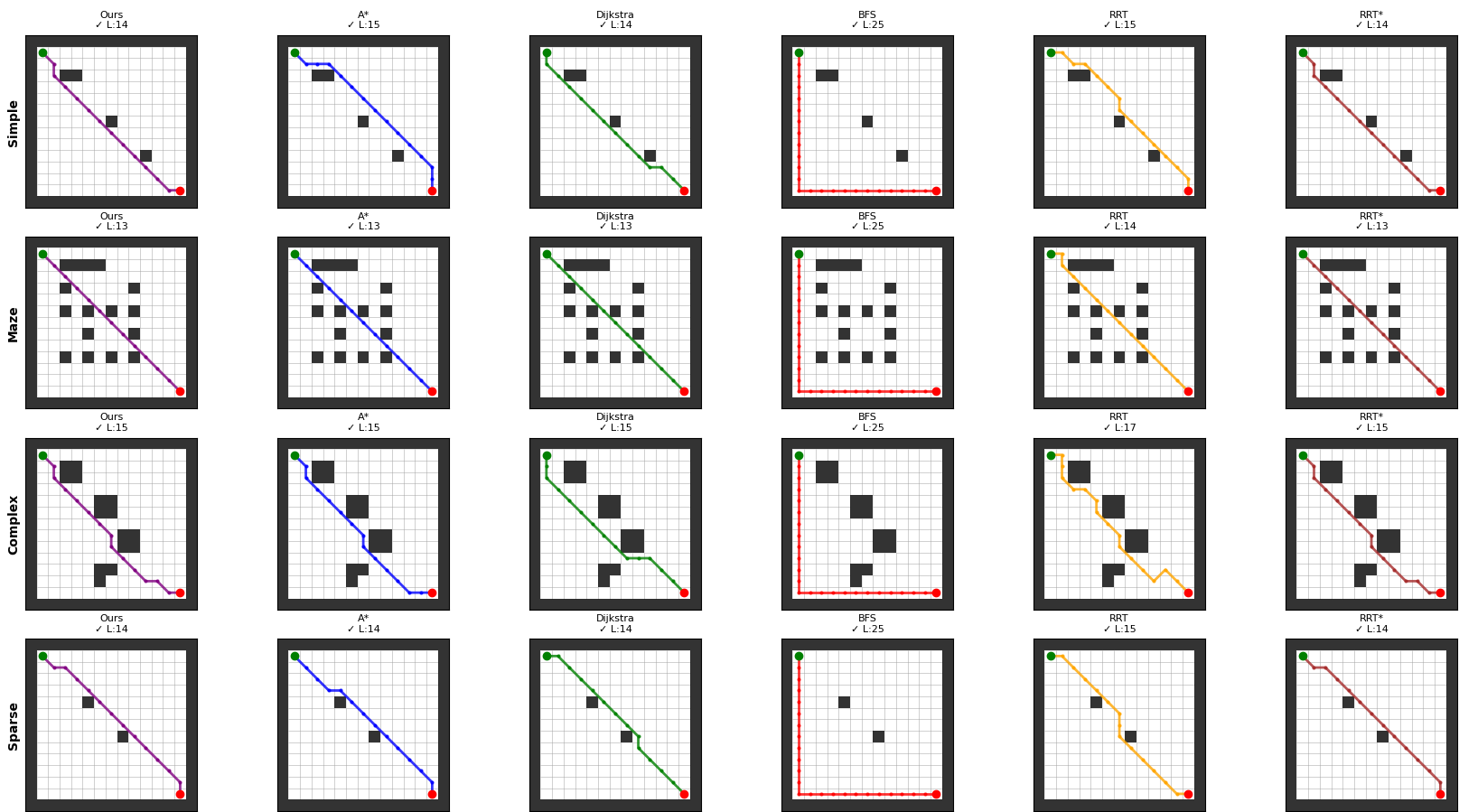}
    \caption{This figure compares our method against five baselines in static environments, demonstrating our approach's superior performance.}
    \label{fig:pdf}
\end{figure*}

\section{Method}

\textbf{GundamQ} framework enhances reinforcement learning in dynamic environments by integrating hierarchical feature extraction, adaptive exploration, and a structured reward for path efficiency, safety, and smoothness. A modified Dueling network outputs optimized actions, improving decision-making while maintaining computational efficiency.

\subsection{Spatiotemporal Feature Extraction}
GundamQ employs spatial attention and multi-scale LSTM \cite{b14} for efficient spatiotemporal feature extraction. The spatial attention addresses the uneven importance of visual information across spatial dimensions. Traditional convolution treats all image regions equally, while practical decision-making relies only on features from the critical areas. We employ a lightweight attention mechanism:

\begin{equation}
A = \sigma(W_c * F)
\end{equation}

\begin{equation}
F_{\text{att}} = A \otimes F
\end{equation}

where \(W_c\) denotes a learnable \(1 \times 1\) convolution kernel, \(\sigma\) is the Sigmoid activation function, and \(\otimes\) denotes element-wise multiplication. This approach autonomously enhances representations of critical regions while suppressing non-essential areas, thereby improving the signal-to-noise ratio of feature representations with minimal computational overhead.

The multi-scale LSTM architecture is motivated by the observation that environmental states exhibit hierarchical temporal dynamics. A short-term LSTM branch captures immediate temporal patterns:

\begin{equation}
h_t^{\text{short}} = \text{LSTM}_{\text{short}}(x_t, h_{t-1}^{\text{short}})
\end{equation}

Simultaneously, a long-term LSTM branch captures macro-level temporal trends by applying downsampling operations to the input sequence:

\begin{equation}
h_t^{\text{long}} = \text{LSTM}_{\text{long}}(\text{DS}(x_t), h_{t-1}^{\text{long}})
\end{equation}

The outputs from both branches are temporally aligned and fused using a multi-head attention mechanism:

\begin{equation}
\text{Attention}(Q, K, V) = \text{softmax}\left(\frac{QK^T}{d}\right)V
\end{equation}

This dual-branch architecture enables rapid response to sudden environmental changes while preserving long-term temporal dependencies.

During feature fusion and decision-making, the model employs an enhanced dueling architecture augmented with learnable action-weight networks:

\begin{equation}
w(a \mid s) = \text{softmax}(\text{MLP}(h_{\text{LSTM}}))
\end{equation}

This design enables dynamic adjustment of action advantage weights during Q-value computation, thereby enhancing policy flexibility. Multiple engineering optimizations, including memory access optimization and parallel computation design, ensure computational efficiency while preserving high modeling fidelity in real-time settings.

\subsection{Adaptive exploration-exploitation Balancing Strategy}

Traditional fixed-parameter exploration strategies struggle to adapt to evolving policies during the training process. To address this, \textbf{GundamQ} introduces a multidimensional adaptive exploration–exploitation strategy, combining dynamic parameter adjustment with intelligent sampling to achieve progressive policy optimization.

GundamQ employs an exponentially decaying $\varepsilon$-greedy strategy to balance exploration and exploitation:
\begin{equation}
\varepsilon_t = \varepsilon_{\min} + (\varepsilon_{\max} - \varepsilon_{\min}) \cdot \exp(-\lambda t),
\end{equation}
where $\varepsilon_{\min}$ and $\varepsilon_{\max}$ denote the lower bound and initial value of the exploration rate, and $\lambda$ controls the decay. This ensures sufficient exploration during early training and a gradual shift to exploitation as training progresses.

The experience replay mechanism is further enhanced by incorporating temporal correlations in addition to TD-error:
\begin{equation}
p_i = |\delta_i| + \alpha \cdot \mathcal{S}(h_t, h_{t+1}),
\end{equation}
where $\delta_i$ denotes the TD-error, $\mathcal{S}(\cdot)$ is a similarity measure based on LSTM hidden states, and $\alpha$ is a scaling factor. This approach prioritizes transitions with strong temporal continuity, thereby improving experience data utilization efficiency.

Exploration intensity is additionally controlled through a temperature scheduling mechanism:
\begin{equation}
T_{t+1} = \gamma T_t,
\end{equation}
where $\gamma$ is a decay coefficient, allowing the policy to gradually shift from high stochasticity to a more deterministic behavior. The temperature parameter regulates exploration randomness, while $\varepsilon$ controls exploration frequency, jointly balancing exploration breadth and exploitation depth.

At the policy optimization level, GundamQ employs a weighted dueling network architecture:
\begin{equation}
Q(s,a) = V(s) + \phi(a|s) A(s,a),
\end{equation}
where $\phi(a|s)$ is a normalized weight generated by an independent network via softmax to model the action-space probability distribution. This design complements the exploration mechanisms and enhances value estimation accuracy in already explored regions.

\subsection{Hierarchical reward function}
The proposed hierarchical reward function is defined as a composite expression:

\begin{equation}
R_t = 
\underbrace{\omega_p \cdot \eta \cdot (d_{t-1} - d_t)}_{\text{distance difference}}
+ 
\underbrace{\omega_c \cdot R_c}_{\text{collision term}}
- 
\underbrace{\omega_s \cdot \beta \cdot \|\Delta^2 a_t\|}_{\text{smoothness term}}
\end{equation}

where:  

\begin{itemize}
    \item $\omega_p, \omega_c = 0.3, \omega_s$ denote the weighting coefficients for the sub-objectives of path progression, collision avoidance penalty, and motion smoothness, respectively, subject to the constraint $\omega_p + \omega_c + \omega_s = 1$.
    \item $\eta$ and $\beta$ are tunable parameters controlling the intensity of path progression incentives and the penalty for motion discontinuities.
    \item $d_t$ represents the Manhattan distance between the agent and the goal at time step $t$.
    \item $R_c$ is the collision penalty term, taking negative values when collisions occur.
    \item $\Delta^2 a_t = a_t - 2 a_{t-1} + a_{t-2}$ is the second-order difference of the agent's acceleration, quantifying abrupt changes in motion.
\end{itemize}
This reward combines distance guidance, collision penalties, and motion smoothness, enabling the agent to learn strategies balancing efficiency, safety, and smoothness. Parameters $\eta$ and $\beta$ can be tuned to accelerate convergence or enhance trajectory smoothness.

\section{Experiment}
We extensively evaluate the proposed method on static and dynamic path planning tasks.
\subsection{ Static Path Planning Performance Evaluation}
\textbf{Experimental setup.}
We validate robustness across four maps: \textit{maze}, \textit{simple}, \textit{complex}, \textit{spiral}. Exploration uses adaptive $\varepsilon$-greedy ($\varepsilon$: $0.9 \rightarrow 0.1$) and Boltzmann exploration. Rewards: goal ($+200$), collision ($-5$), movement ($+0.2$). Key hyperparameters: learning rate $1 \times 10^{-3}$, $\gamma = 0.99$, batch size $64$, LSTM length $10$.

The \textbf{success rate} based on path completion is defined as:

\begin{equation}
SR = \min(1, \frac{L_{valid}}{L_{optimal}})
\end{equation}

where $L_{valid}$ represents the length of the actual valid path traversed by the robot, and $L_{optimal}$ denotes the optimal path length from start to goal.

\textbf{Average Path Length (Avg Len.)}: Mean distance of successfully generated trajectories, indicating spatial efficiency of navigation solutions.

\textbf{Path Length Ratio (Ratio)}: Normalized ratio of computed path to optimal path length, measuring proximity to optimality.

\textbf{Search Time (Time)}: Computational latency for trajectory generation in milliseconds, reflecting algorithmic efficiency and real-time capability.

\textbf{Smoothness (Smooth.)}: Trajectory quality measured by directional changes, with lower values indicating better kinematic feasibility.

\textbf{Standard Deviation (Std Dev)}: Statistical dispersion of success rates across trials, quantifying algorithmic consistency and robustness.

\begin{figure}[t]
    \centering
    \includegraphics[width=0.8\columnwidth]{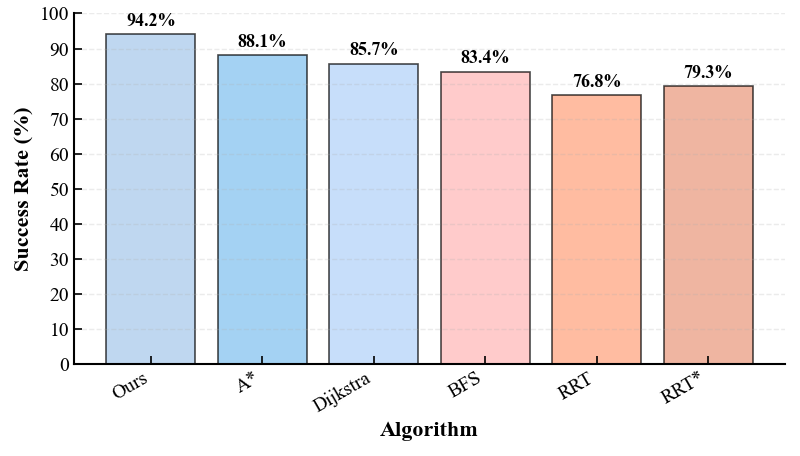}
    \caption{Algorithm Success Rate Comparison}
    \label{fig:success_rate}
\end{figure}
\begin{table}[t]
\centering
\scalebox{1}{
\begin{tabular}{l|cccccc}
\hline
\textbf{Alg.} & \textbf{SR} & \textbf{Len.} & \textbf{Ratio} & \textbf{Time} & \textbf{Smooth.} & \textbf{Std} \\
& \textbf{(\%)} & & & \textbf{(ms)} & & \textbf{Dev} \\
\hline
\textbf{Gundam} & \cellcolor{gray!20}\textbf{94.2} & \cellcolor{gray!20}\textbf{21.6} & \cellcolor{gray!20}\textbf{1.1} & \cellcolor{gray!20}\textbf{0.2} & \cellcolor{gray!20}\textbf{2.1} & \cellcolor{gray!20}\textbf{$\pm$3.2} \\
\hline
A* & 88.1 & 20.8 & 1.0 & 1.2 & 1.8 & $\pm$4.1 \\
\hline
Dijkstra & 85.7 & 20.5 & 1.0 & 2.2 & 1.6 & $\pm$5.2 \\
\hline
BFS & 83.4 & 22.3 & 1.1 & 3.5 & 2.4 & $\pm$6.1 \\
\hline
RRT & 76.8 & 28.4 & 1.4 & 8.7 & 5.2 & $\pm$12.3 \\
\hline
RRT* & 79.3 & 26.7 & 1.4 & 15.6 & 4.8 & $\pm$11.1 \\
\hline
\end{tabular}
}
\caption{Comparison of pathfinding algorithms performance metrics}
\label{tab:pathfinding_comparison}
\end{table}

\textbf{Baseline.}
We select five representative algorithms as baselines: A* \cite{b15}, Dijkstra \cite{b16}, BFS \cite{b17}, RRT \cite{b18}, and RRT* \cite{b19}, representing the three major paradigms of graph search, sampling-based planning, and learning-based planning, respectively. All algorithms operate on identical grid environments. RRT-based methods are averaged over 5 runs, while deterministic algorithms require only single executions.

\textbf{Result.}
GundamQ achieves superior performance on static path planning tasks compared to baseline methods. Regarding success rate, GundamQ attains 94.2\%, significantly outperforming A* (88.1\%), Dijkstra (85.7\%), BFS (83.4\%), RRT (76.8\%), and RRT* (79.3\%). For path quality, GundamQ exhibits an average path length ratio of 1.08, only 8\% longer than the optimal path, substantially superior to the RRT family's 1.35-1.42. With an inference time of 0.23ms, 5.4× faster than A* (1.24ms), it demonstrates the real-time advantages of end-to-end learning.

\subsection{Ablation Experiment}

To evaluate the individual contributions of key modules within the GundamQ framework, we conduct comprehensive ablation studies. Each variant is constructed by systematically disabling specific components while maintaining all other parameters constant. The analysis focuses exclusively on variants of our proposed method.

\textbf{Experimental Setup:} We recorded three core metrics: path planning steps, execution time (ms), and success rate (\%).

\textbf{Result.}
The ablation study results demonstrate the critical importance of each framework component. The removal of the spatial attention mechanism (A1) resulted in performance degradation: planning time increased from 0.23ms to 0.41ms, path length from 18.4 to 25.7 steps, and success rate from 94.2\% to 86.8\%. Eliminating the exploration balancing strategy (A3) caused planning time to rise to 0.35ms, path steps to 31.2, and success rate to drop to 78.9\%. Similarly, removing the multi-scale LSTM component (A2) resulted in planning time of 0.38ms, path length of 28.9 steps, and success rate of 81.5\%. These results confirm that each module makes substantial contributions to overall system performance.

\begin{table}[h]
\centering
\begin{tabular}{l|ccc}
\hline
\textbf{Variant} & \textbf{Steps} & \textbf{Time (ms)} & \textbf{SR (\%)} \\
\hline
\textbf{Full GundamQ} & 18.4 & 0.23 & 94.2 \\
\hline
A1: w/o Spatial Attention & 25.7 & 0.41 & 86.8 \\
\hline
A2: w/o Multi-scale LSTM & 28.9 & 0.38 & 81.5 \\
\hline
A3: w/o Adaptive Exploration & 31.2 & 0.35 & 78.9 \\
\hline
\end{tabular}
\caption{Ablation experiment results showing the impact of removing key components}
\label{tab:ablation_study}
\end{table}

\begin{table}[b]
\centering
\scalebox{1}{
\begin{tabular}{l|cccc}
\hline
\textbf{Alg.} & \textbf{SR} & \textbf{Len.} & \textbf{Time} & \textbf{Ratio} \\
& \textbf{(\%)} & & \textbf{(s)} & \\
\hline
\textbf{GundamQ} & \cellcolor{gray!20}\textbf{87.6} & \cellcolor{gray!20}\textbf{24.6} & \cellcolor{gray!20}\textbf{1.6} & \cellcolor{gray!20}\textbf{1.15} \\
\hline
A* & 72.3 & 31.4 & 2.8 & 1.08 \\
\hline
RRT & 68.9 & 35.2 & 3.4 & 1.42 \\
\hline
Adaptive A* & 75.8 & 29.7 & 2.1 & 1.12 \\
\hline
\end{tabular}
}
\caption{Performance comparison in   Dynamic Environment  Evaluations}
\label{tab:dynamic_comparison}
\end{table}
\begin{table}[b]
\centering
\scalebox{1}{
\begin{tabular}{l|c|ccc}
\hline
\textbf{Complexity} & \textbf{Density} & \textbf{Ours} & \textbf{A*} & \textbf{RRT} \\
& \textbf{(\%)} & \textbf{(\%)} & \textbf{(\%)} & \textbf{(\%)} \\
\hline
Low & 5 & 94.2 & 86.1 & 82.3 \\
\hline
Medium & 15 & 87.6 & 72.3 & 68.9 \\
\hline
High & 25 & 78.4 & 58.7 & 52.1 \\
\hline
Extreme & 35 & 71.9 & 41.2 & 38.6 \\
\hline
\end{tabular}
}
\caption{Success rate comparison across different environment complexities}
\label{tab:complexity_comparison}
\end{table}
\subsection{ Dynamic Environment Evaluation}
This section evaluates the performance of our proposed method in dynamic environments. We augment static grid maps with dynamic mechanisms that randomly alter grid configurations to simulate real-world variable conditions. We conducted comparative analysis against A$^*$, RRT and Adaptive A$^*$ algorithms using success rate metrics from Section A.

\textbf{Experimental setup.}
To validate path planning capabilities in dynamic environments, we introduce dynamic obstacles into static grid environments with time-varying configurations. This setup produced results in Table III. We design three grid environments with varying difficulty levels to evaluate navigation performance across different scenarios.The difficulty is defined as:

\begin{equation}
\text{Density} = \frac{N_{\text{obstacles}}}{N_{\text{total}}} \times 100\%
\end{equation}

where Nobstacles represents the number of obstacle cells and Ntotal represents the total number of grid cells.
\begin{figure}[t]
    \centering
    \includegraphics[width=0.75\linewidth]{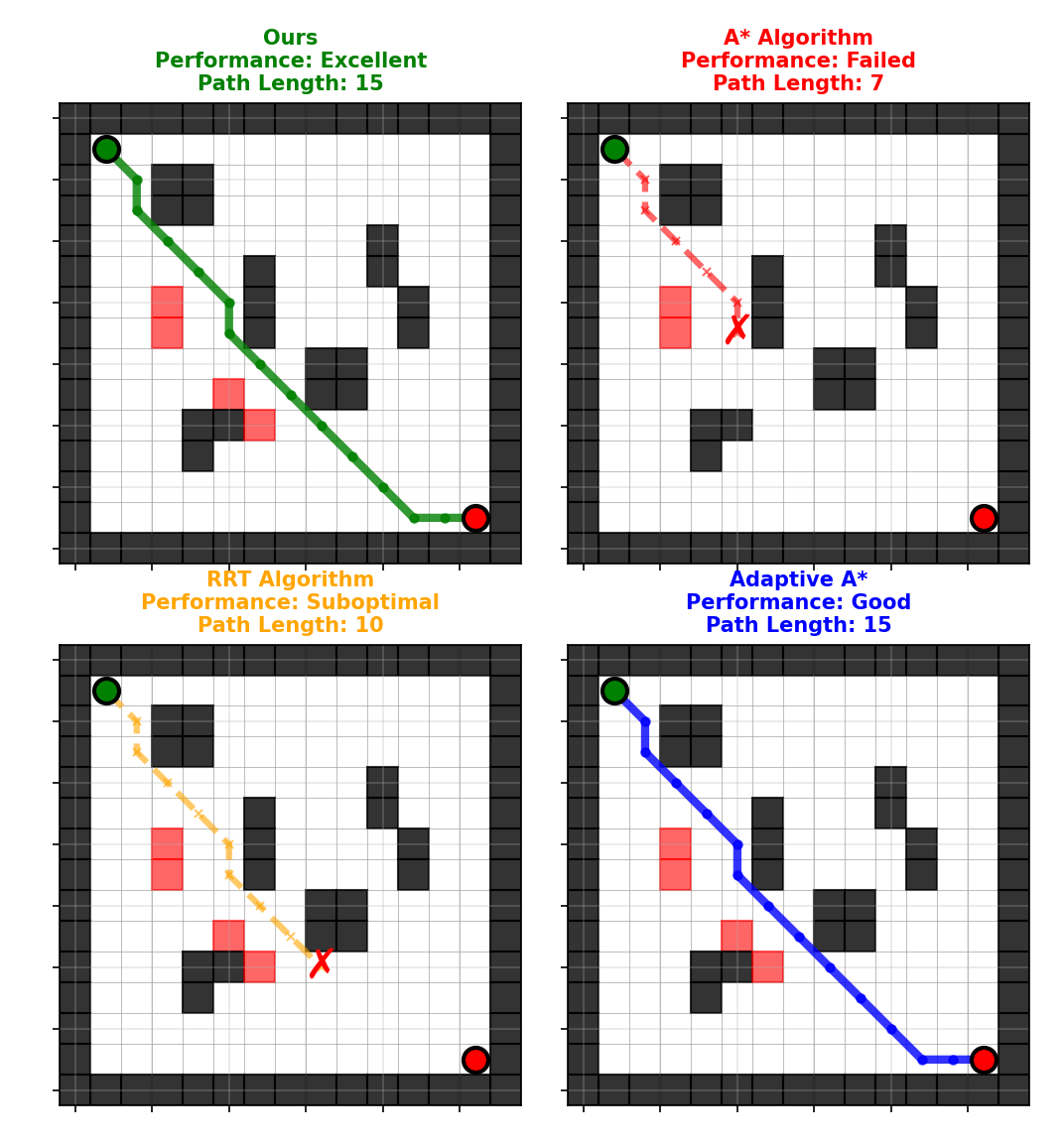} 
    \caption{This figure shows the comparative performance of our method against two baseline approaches in dynamic environments. Red blocks indicate obstacles generated during execution.}
    \label{fig:pdfimage}
\end{figure}

\textbf{Result.}
The proposed spatial attention mechanism and adaptive exploration strategy achieve substantial improvements over baseline methods. Specifically, our approach reduces average path planning steps by 6.8 steps (from 31.4 to 24.6), decreases execution time by 1.2 seconds (from 2.8 to 1.6 seconds), and enhances success rate by 15.3\% (from 72.3\% to 87.6\%). These results demonstrate the effectiveness of the proposed method across diverse dynamic environments.

\section{Conclution and Discussion}

We propose GundamQ, a framework that addresses robot path planning in dynamic and complex environments through multi-scale spatiotemporal temporal representation learning combined with spatial attention mechanisms and dual-branch LSTM architecture.
Our approach has several limitations: experiments are primarily conducted in 2D grid environments, the multi-scale LSTM introduces computational overhead, and the reward function requires manual hyperparameter tuning.
Future research directions include: extending the framework to 3D environments, developing lightweight architectures, and integrating uncertainty quantification for safety-critical applications.


\begin{thebibliography}{00}
\bibitem{b1} Y. Shaoul, I. Mishani, S. Vats, J. Li, and M. Likhachev, ``Multi-Robot Motion Planning with Diffusion Models,'' in \textit{Proceedings of the International Conference on Learning Representations (ICLR)}, 2025.
\bibitem{b2}M. Torne, M. Balsells, Z. Wang, S. Desai, T. Chen, P. Agrawal, and A. Gupta, ``Breadcrumbs to the Goal: Goal-Conditioned Exploration from Human-in-the-Loop Feedback,'' in \textit{Advances in Neural Information Processing Systems 36 (NeurIPS)}, 2023.
\bibitem{b3}Y. Cao, T. Hou, Y. Wang, X. Yi, and G. Sartoretti, ``ARiADNE: A Reinforcement learning approach using Attention-based Deep Networks for Exploration,'' in \textit{Proc. 2023 IEEE Int. Conf. Robotics and Automation (ICRA)}, pp. 10219--10225, 2023.
\bibitem{b4} H.-T. L. Chiang, J. Hsu, M. Fiser, L. Tapia, and A. Faust, ``RL-RRT: Kinodynamic Motion Planning via Learning Reachability Estimators From RL Policies,'' \textit{IEEE Robot. Automat. Lett.}, vol. 4, no. 4, pp. 4298--4305, Oct. 2019.
\bibitem{b5}A. Vashisth, J. Rückin, F. Magistri, C. Stachniss, and M. Popović, ``Deep Reinforcement Learning with Dynamic Graphs for Adaptive Informative Path Planning,'' \textit{IEEE Robot. Automat. Lett.}, vol. 9, no. 9, pp. 7747--7754, 2024.
\bibitem{b6}W. Hu, H. Zhang, and Y. Zhou, ``Hierarchical Deep Deterministic Policy Gradient for Autonomous Maze Navigation of Mobile Robots,'' arXiv preprint arXiv:2508.04994, 2024.
\bibitem{b7}K. Chua, R. Calandra, R. McAllister, and S. Levine, ``Deep Reinforcement Learning in a Handful of Trials using Probabilistic Dynamics Models,'' in \textit{Advances in Neural Information Processing Systems 31 (NeurIPS)}, pp. 4754--4765, 2018.

\bibitem{b8}A. Kumar, A. Zhou, G. Tucker, and S. Levine, ``Conservative Q-Learning for Offline Reinforcement Learning,'' in \textit{Advances in Neural Information Processing Systems 33 (NeurIPS)}, pp. 9592--9602, 2020.
\bibitem{b9}M. Gök, E. Esen, A. Gök, G. Yazici, and T. Koyun, ``Dynamic path planning via Dueling Double Deep Q-Network (D3QN) with prioritized experience replay,'' \textit{Applied Soft Computing}, vol. 154, p. 111334, Mar. 2024.
\bibitem{b10}J. Chiun, S. Zhang, Y. Wang, Y. Cao, and G. Sartoretti, ``MARVEL: Multi-Agent Reinforcement Learning for constrained field-of-View multi-robot Exploration in Large-scale environments,'' in \textit{Proceedings of the IEEE International Conference on Robotics and Automation (ICRA)}, Atlanta, GA, USA, May 19-23, 2025.
\bibitem{b11}I. El Shar and D. R. Jiang, ``Weakly Coupled Deep Q-Networks,'' in \textit{Advances in Neural Information Processing Systems 36 (NeurIPS)}, 2023.
\bibitem{b12}R. Rafailov, A. Sharma, E. Mitchell, S. Ermon, C. D. Manning, and C. Finn, ``Direct Preference Optimization: Your Language Model is Secretly a Reward Model,'' in \textit{Advances in Neural Information Processing Systems 36 (NeurIPS)}, pp. 24696--24723, 2023.
\bibitem{b13}S. Liu, ``An Evaluation of DDPG, TD3, SAC, and PPO: Deep Reinforcement Learning Algorithms for Controlling Continuous System,'' in \textit{Proceedings of the 2023 International Conference on Data Science, Advanced Algorithm and Intelligent Computing (DAI 2023)}, pp. 15--24, 2024.
\bibitem{b14}J. Chung, S. Ahn, and Y. Bengio, ``Hierarchical Multiscale Recurrent Neural Networks,'' in \textit{5th International Conference on Learning Representations (ICLR)}, 2017.
\bibitem{b15}L. Zhang, X. Wang, and Y. Chen, ``Enhancing a star algorithm for robot path planning,'' in \textit{Proceedings of the 2023 International Conference on Machine Learning and Automation (ICML)}, pp. 245--250, 2023.
\bibitem{b16}B. Haeupler, R. Hladík, V. Rozhoň, R. E. Tarjan, and J. Tětek, ``Universal Optimality of Dijkstra via Beyond-Worst-Case Heaps,'' in \textit{Proceedings of the 2024 IEEE 65th Annual Symposium on Foundations of Computer Science (FOCS)}, pp. 2099--2130, 2024.
\bibitem{b17}B. Haeupler, R. Hladík, V. Rozhoň, R. E. Tarjan, and J. Tětek, ``Universal Optimality of Dijkstra via Beyond-Worst-Case Heaps,'' in \textit{Proceedings of the 2024 IEEE 65th Annual Symposium on Foundations of Computer Science (FOCS)}, pp. 2099--2130, 2024.
\bibitem{b18}A. J. LaValle, B. Sakcak, and S. M. LaValle, ``Bang-bang Boosting of RRTs,'' in \textit{Proceedings of the 2023 IEEE/RSJ International Conference on Intelligent Robots and Systems (IROS)}, pp. 2869--2876, 2023.
\bibitem{b19}S. Karaman and E. Frazzoli, ``Incremental Sampling-based Algorithms for Optimal Motion Planning,'' in \textit{Proceedings of Robotics: Science and Systems (RSS)}, Zaragoza, Spain, 2010.


\end{thebibliography}
\end{document}